\documentclass[10pt,twocolumn,letterpaper]{article}

\usepackage{cvpr}              

\usepackage{graphicx}
\usepackage{amsmath}
\usepackage{amssymb}
\usepackage{booktabs}
\usepackage{bm}

\usepackage[pagebackref,breaklinks,colorlinks]{hyperref}

\usepackage[capitalize]{cleveref}
\crefname{section}{Sec.}{Secs.}
\Crefname{section}{Section}{Sections}
\Crefname{table}{Table}{Tables}
\crefname{table}{Tab.}{Tabs.}

\def\cvprPaperID{10630} 
\def\confName{CVPR}
\def\confYear{2023}

\begin{document}

\title{Frequency Decomposition to Tap the Potential of Single Domain for Generalization}

\author{
Qingyue Yang, Hongjing Niu, Pengfei Xia, Wei Zhang, Bin Li 
\thanks{Corresponding Author}\\
University of Science and Technology of China\\
{\tt\small \{yangqingyue, sasori, xpengfei, zw1996\}@mail.ustc.edu.cn, binli@ustc.edu.cn}}


\maketitle

\begin{abstract} 
   Domain generalization (DG), aiming at models able to work on multiple unseen domains, is a must-have characteristic of general artificial intelligence. DG based on single source domain training data is more challenging due to the lack of comparable information to help identify domain invariant features. In this paper, it is determined that the domain invariant features could be contained in the single source domain training samples, then the task is to find proper ways to extract such domain invariant features from the single source domain samples. An assumption is made that the domain invariant features are closely related to the frequency. Then, a new method that learns through multiple frequency domains is proposed. The key idea is, dividing the frequency domain of each original image into multiple subdomains, and learning features in the subdomain by a designed two branches network. In this way, the model is enforced to learn features from more samples of the specifically limited spectrum, which increases the possibility of obtaining the domain invariant features that might have previously been defiladed by easily learned features. Extensive experimental investigation reveals that 1) frequency decomposition can help the model learn features that are difficult to learn. 2) the proposed method outperforms the state-of-the-art methods of single-source domain generalization.

\end{abstract}

\section{Introduction} 
\label{sec:intro}

\begin{figure}
	\centering          
	\includegraphics[scale=1]{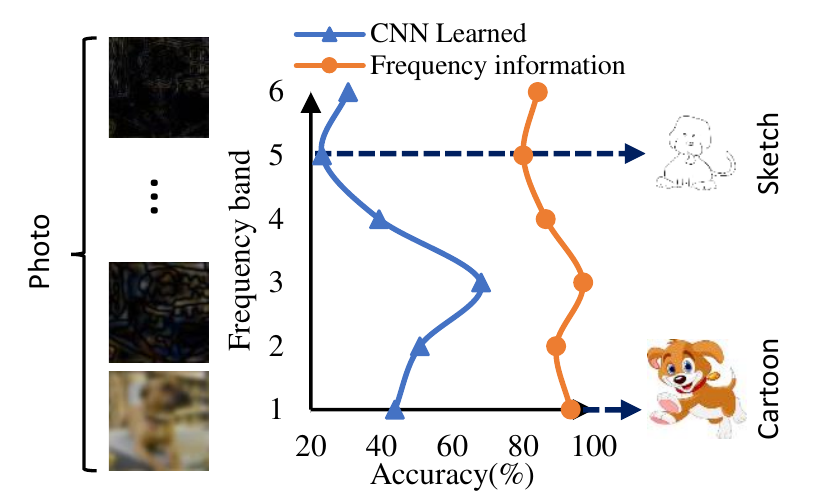}  
\caption{Accuracy of different frequency slices of photo domain image. The orange line with round dots is the accuracy of training with each frequency band and evaluating on itself. It means the classification information of each frequency band in the photo domain. The blue line with triangle dots is the accuracy of training with original images and evaluating on each frequency band. There is a large gap between the two lines which we want to fill. Different domains have different main frequency bands, \eg cartoon images have the most information in the frequency band 1, and sketch images are concentrated in the frequency band 5. So we want to learn all frequency bands well.}
\label{fig:sketch}
\end{figure}
Deep learning has shown excellent performance in various fields \cite{szegedy2015going, simonyan2014very, krizhevsky2017imagenet, he2016deep, huang2017densely}. 
However, basing on the assumption that training samples and testing samples are from independent and identically distribution is a limitation on generalization capability. 
In real-world scenarios, the assumption is hardly to be satisfied because of lighting, background, and other unpredictable factors.
At this time, the performance of a model will be greatly reduced \cite{ben2010theory}. 
Therefore, domain generalization capability is essential for machine learning models. 


Domain generalization (DG) is a classical machine learning task, which has also attracted much attention in recent years. 
The most common setting is multi-source DG tasks, in which a model is trained with samples from multiple source domains and tested on unseen target domains.
The domains follow different distributions. 
For example, in image classification tasks, datasets with different domains may come from different acquisition methods or have different image styles \cite{ben2010theory}. 

An intuitive idea for DG tasks is to learn the common features of the source domains.
For example, some approaches \cite{motiian2017unified, wang2018visual} minimize the distance between samples of the same categories from different domains. 
However, the collection and annotation of data in multiple domains cost too much in many scenarios. So the single domain generalization (SDG) approaches are required.

In the case of single domain generalization, in which only one source domain is available, it is more challenging for a model to learn enough knowledge for generalizing to multiple domains. 
Due to the lack of comparable information to help identify domain invariant features, the single domain generalization task is more challenging.
The method of finding commonality between source domains, such as invariant risk minimization\cite{arjovsky2019invariant}, will no longer be effective.
There are some interesting studies on SDG tasks.
For example, some methods \cite{li2021progressive, cugu2022attention, qiao2021uncertainty, wang2021learning} use data augmentation to enhance model generalization. Those methods add noise to the image or generate new style images to simulate the domain changes. 
However, the facticity and sufficiency of those modifications are hard to be guaranteed, which limits their wide application.
In addition to introducing new perturbations, further mining the effective information in the samples is also a feasible strategy. 
Thus, we proposed a method that decomposes the image and learns all information that does not use pseudo-domain information, unlike former methods. 
The motivation behind this idea is that the domain invariant representation might be buried in the images but challenging to learn. 
In many cases, deep learning models may not be able to learn all the features in the data that are helpful to the task.
For example, deep learning models may only learn shortcuts \cite{geirhos2020shortcut}, which are decision rules that perform well on standard benchmarks but fail to transfer to real-world scenarios.
We assume that the poor performance of deep learning models on SDG tasks is partly due to the failure to fully learn all the effective features in the given dataset.
Therefore, our approach decomposes the image, so the hidden information can be exposed.
Through the decomposition, the domain invariant features, which might previously be defiladed by easily learned features, could be learned.

We use frequency domain decomposition to decompose the image while preserving the effective features of each part as much as possible.
The frequency domain is a widely used aspect for decomposing images. Many frequency methods have been applied to DG tasks. 
Some \cite{xu2021fourier} found that the high-frequency information of images depicts object edge structure, which is naturally consistent across different domains. So many works tried to divide frequency into a domain related and invariant from different perspectives and mix the related ones.
Jeon \etal~\cite{jeon2021feature} divides the image into two parts with frequency and mixed the low-frequency parts, which are related to domain style, to achieve the effect of domain generalization. Chen \etal~\cite{chen2021amplitude} decomposes frequency in terms of phase and amplitude. 
Unlike defining some frequencies as domain invariant frequencies, our method decomposes samples into many frequency bands and aims at learning them all.

To verify the feasibility of frequency domain decomposition, we tested the performance of classification tasks using different frequency bands. 
And the result is shown in \cref{fig:sketch}. 
On the one hand, it can be seen that the model trained on each frequency band of images in the Photo domain (orange line) achieves over 80\% accuracy, which indicates that each frequency band contains a considerable amount of effective features. 
On the other hand, the model trained on original images in the Photo domain (blue line) does not perform well when evaluated on any frequency bands.
Combining the two observations, we assume that each frequency band contains effective information of an image, but it is often difficult to be fully learned by the model.
More details and discussions about this experiment are in the Experiments section (\cref{sec:pagestyle}).

We proposed an approach to fully learn different frequency components based on the assumption.
Specifically, we designed a dual-branch model, and the two branches receive complementary frequency bands as input. And a similarity loss is used to shorten the distance between complementary frequency bands. The division of frequency bands is diversified to ensure sufficient learning of each frequency component.

Our contributions can be summarized as follows: 
\begin{itemize}
    \item  We found that each frequency component contains effective information that cannot be ignored, but not all of them are learned by deep learning models. 
    \item Based on the above observation, we propose a new domain generalization method, which can better learn the effective features contained in each frequency band.
    And the proposed method has achieved state-of-the-art accuracy.
\end{itemize}


\section{Related Work}
\label{sec:relatedwork}

\textbf{Domain generalization}~\cite{shen2021towards, volpi2018generalizing, zhou2021domain} aims to learn a model on source domains that can perform well on unseen domains.
Many methods are proposed to solve domain generalization tasks.
Some approaches \cite{ghifary2016scatter, jin2020feature} align source domain distributions by minimizing moments of transformed features between source domains, to learn domain-invariant representation.
Li \etal~\cite{li2018learning} first applied meta-learning to DG and many people \cite{balaji2018metareg, zhao2021learning} followed this work. In these methods, source domains are divided into non-overlapping meta-source and meta-target domains to simulate domain shift.
Many methods use data augmentation, such as image transformation\cite{volpi2019addressing},  random augmentation\cite{xu2020robust}, and feature-based augmentation\cite{zhou2021mixstyle} to solve DG tasks. While Volpi \etal~\cite{volpi2018generalizing} uses adversarial gradients obtained from the classifier to perturb the input images, so the coverage of the training domain is expanded.

\textbf{Single domain generalization} is a more challenging domain generalization task. The task setting is similar to normal DG (multi-source DG) except that only one source domain is available in training the model. 
So it is no longer possible to obtain domain invariant representation by finding the commonality between source domains.
To solve this difficult task, former proposed works put the emphasis on meta-learning and introducing disturbance.
Wang \etal~\cite{wan2022meta} decomposes image features into meta features, so as to encode an image without domain information. Qiao and Peng \cite{qiao2021uncertainty} expose the model to domain shift during training via meta-learning, and the synthesis of out-of-domain data is guided by uncertainty assessment.

Introducing disturbance, such as adversarial training and expanding the coverage of the training domain, aims to generate out-of-domain data. By training with these samples, which simulate the changed domains, the model can learn domain invariant features and enhance the generalization ability.
Qiao \etal~\cite{qiao2020learning} and Fan \etal~\cite{fan2021adversarially} use adversarial training to create fictitious but challenging groups, from which the model can be learned and promoted under the theoretical guarantee. 
Research based on data augmentation adds noise to an image or its feature to let the model learn domain invariant features. 
Cugu \etal~\cite{cugu2022attention} uses multiple visual corruptions to alter the training images to simulate new domains. 
Li \etal~\cite{li2021progressive} and Wang \etal~\cite{wang2021learning} generate the extended domains by using comparative learning or mutual information to guarantee the safety and effectiveness of the extended domain. 

The implicit assumption of these methods is that the introduced invariance is effective for domain generalization. 
Unlike those data augmentation methods, our approach adds nothing to the image but decomposes samples into slices. By fully learning them, the model can learn domain invariant features.

\textbf{Frequency domain} options are incorporated into deep learning methods for enhancing the robustness and generalization capability of the model. 
Many works focused on image frequency processing.
Jeon \etal~\cite{jeon2021feature} changes the low frequency component of the image to affect the image style and keeps the high frequency component unchanged to maintain the shape information. By using this data augmentation, the model aims to learn domain invariant features.
Chen \etal~\cite{chen2021amplitude} supposes that humans recognize images more through phase information. So they propose a method to keep the phase of the image unchanged and add disturbance to the amplitude to increase the robustness of CNN (Convolutional Neural Networks).
Guo \etal~\cite{guo2018low} uses low-frequency perturbations on the image for adversarial attacks.

On the other hand, some methods convert features to frequency domain for processing.
Lin \etal~\cite{lin2022deep} transforms the feature maps of different network layers into the frequency domain. Then a mask is generated to enhance the domain invariant frequency components and suppress the components that are not conducive to generalization.
Guo and Ouyang \cite{guo2020robust} learn an effective frequency range for the features of each convolution layer to improve the convergence and robustness of CNN.
Our method decomposes images from the frequency domain and aims at learning them all rather than supposing that there are frequency components with good generalization.

\section{Method} 
\label{sec:method}

\begin{figure*}[htbp]
\begin{subfigure}{0.41\linewidth}
    \centering          
	\includegraphics[scale=0.98]{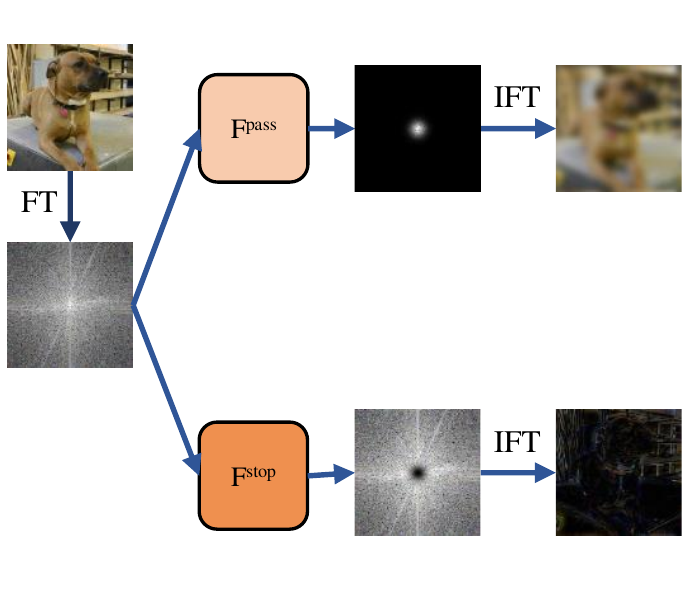}   
	\caption{ }
	\label{subfig:main_a}
\end{subfigure}
\begin{subfigure}{0.59\linewidth}
	\centering      
	\includegraphics[scale=1]{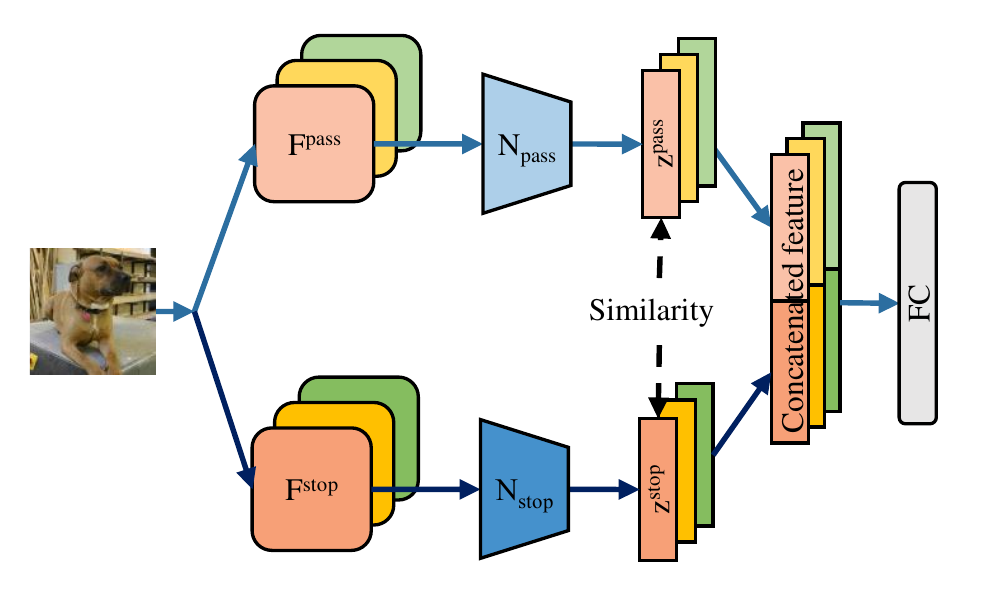}   
	\caption{ }
	\label{subfig:main_b}
\end{subfigure}
\caption{The overall framework of the proposed method. (a) Illustration of the frequency slices of the image. The image is divided into two slices in the frequency domain by a pair of complementary filters. (b) The training process. The image is processed by a pair of filters for each frequency band, then input to two neural networks. The similarity between the two features extracted from the image is calculated.}
\label{fig:main_method}
\end{figure*}

This section presents our approach for single domain generalization tasks through learning features of all efficient frequency components in the source domain.
We first define the problem in Section \cref{subsec:definition} and then illustrate the image frequency dividing method in Section \cref{subsec:frequency}. Finally, we propose the framework of our approach in Section \cref{subsec:framework}.

\subsection{Problem Definition}
\label{subsec:definition}

The single source domain generalization task is a machine learning task with a wide range of application scenarios and challenges.
In the training phase, only data from a single source domain is available. 
And the test scenario is composed of several target domains that have not been seen in the training phase.

Different domains share the same category space for image classification tasks, but the image style or acquisition way is inconsistent.
To solve the SDG problem, extracting common features between source and target domains is very necessary.
While the target domain is unknown, the source domain is the only possible source of common domain features
With only one source domain available, it is hard to decide which features may perform better in generalization. Therefore, a feasible strategy is to extract as many features as possible that are helpful for classification.
The frequency domain is an effective tool for decomposing images. We found that each frequency component contains a considerable amount of effective features for image classification, but it is often difficult to be fully learned by the model.

Therefore, we propose a method that can learn the information of different frequency components to solve the single source domain generalization task.

\subsection{Frequency Slice of Image}
\label{subsec:frequency}
As shown in \cref{subfig:main_a}, we apply Fourier transformation (FT) to the three RGB channels of the image $\bm{x}$ to transform the image from the spatial domain to the frequency domain. Then we use a Gaussian bandpass filter $F^{pass}$ to take out part of the frequency band of the image. Meanwhile, there is a complementary bandstop filter $F^{stop}$, which stop frequency is consistent with the pass frequency of the bandpass filter. For this pair of bandpass and bandstop filters, 
\begin{equation}
  H^{pass}(\bm{x})+H^{stop}(\bm{x})=\bm{1}, 
  \label{eq:passstop}
\end{equation}
where $H^{pass}(\cdot)$ denotes the frequency response of $F^{pass}(\cdot)$ and $H^{stop}(\cdot)$ denotes the frequency response of $F^{stop}(\cdot)$. 
Finally, we use inverse Fourier transformation (IFT) to transform the frequency components back to the spatial domain. That is,
\begin{equation}
  \bm{x}^{pass}_i = IFT( FT( \boldsymbol{x}) \circ F^{pass}_i)
  \label{eq:FTpass}
\end{equation}
\begin{equation}
  \bm{x}^{stop}_i = IFT( FT( \boldsymbol{x}) \circ F^{stop}_i)
  \label{eq:FTstop}
\end{equation}
where $ \boldsymbol{x}^{pass}_i$ and $ \bm{x}^{stop}_i$ denote the frequency slice of the image, $\circ $ denotes Hadamard product.

The two filters decompose the image into two complementary parts about the pass frequency band, which contain specific image information about this frequency component. 
The pair of filters helps the model extract features from specific frequency bands. 

\subsection{Overall Framework}
\label{subsec:framework}

Our training process is shown in \cref{subfig:main_b}. 
Firstly, we apply FT to the input image to get the spectrogram. 
Then the model has two branches, the upper \textbf{pass branch} and the lower \textbf{stop branch}, which means the filters in each branch are bandpass filters and bandstop filters.

In the pass branch, $K$ Gaussian bandpass filters $F^{pass}_{i=1\sim K}$ decompose the image into $K$ pieces in the frequency domain. The pass frequencies of $F^{pass}_{i=1\sim K}$ satisfy
\begin{equation}
  \sum_{i = 1}^{K}{H^{pass}_i(\bm{x})} = \bm{1}
  \label{eq:passband}
\end{equation}
where $H^{pass}_i(\cdot)$ denotes the frequency response of $F^{pass}_{i}$. 
Then the frequency pieces are transformed back to the spatial domain and become frequency slices $\bm{x}^{pass}_{i=1 \sim K}$ as the \cref{subsec:frequency}.
After that, the frequency slices $\bm{x}^{pass}_i$ are send to a neural network $ N_{pass}(\cdot)$ to extract features $\bm{z}_i^{pass}$.
The frequency slices only have the information of their own frequency bands and let the network learn information of every frequency band precisely. In this way, the network can fully learn the information of the image, including the domain invariant information. 

In the stop branch, a series of Gaussian bandstop filters $F^{stop}_{i=1 \sim K}$ are used. Their stop frequency is consistent with the pass frequency of the pass branch as the \cref{eq:passstop} illustrated. The stop frequencies of $F^{stop}_{i}$ satisfy
\begin{equation}
  \frac{1}{K-1}\sum_{i = 1}^{K}{H^{stop}_i(\bm{x})} = \bm{1}, K \geq 2, 
  \label{eq:stopband}
\end{equation}
where $H^{stop}_i(\cdot)$ denotes the frequency response of $F^{stop}_{i}$.
After the filters $F^{stop}_{i}$, the frequency slices $\bm{x}^{stop}_i$ are send to a neural network $ N_{stop}(\cdot)$ as the pass branch and become features $\bm{z}_i^{stop}$.
Except for the frequency slices, the original image is also sent to the two neural networks to provide information from the whole frequency band aspect.
Since the two branches aim to extract effective features of the same object from different frequency bands, the features should be similar to some extent.
So the consistency loss $L_{cons}(\bm{z}_i^{pass}, \bm{z}_i^{stop})$ is calculated for the features. We use cosine similarity loss as the $L_{cons}$. This ensures that the two networks can extract effective features which exist in a pair of non-overlapping frequency bands.
 
Finally, each pair of features $\bm{z}_i^{pass}$ and $\bm{z}_i^{stop}$ are concatenated together and input to a fully connected layer $FC(\cdot)$ to obtain the classification results $\widehat{y}_i$. Between the true label $y$ and every predicted label $ \widehat{y}_i$, classify loss $L_{cls}(y, \widehat{y}_i)$ is calculated to ensure the correct classification. We use cross-entropy loss as the $L_{cls}$.
The whole loss function is
\begin{equation}
  L = \sum_{i=1}^K{L_{cls}(y, \widehat{y}_i) + \alpha L_{cons}(\bm{z}_i^{pass}, \bm{z}_i^{stop}) }
  \label{eq:whole_loss}
\end{equation}
where $\alpha$ is the trade-off hyper-parameter.

In test phase, the test image is directly input to the two networks $N_{pass}$ and $N_{stop}$ without filters. Then features are connected together and input to the liner layer to get the predicted class label.

\section{Experiment} 
\label{sec:pagestyle}
\subsection{Experimental Setup}

\textbf{Datasets. }
To evaluate the proposed method, we conduct experiments over PACS\cite{li2017deeper}, which is a widely used SDG benchmark. Since the styles vary greatly between domains, it is a challenging benchmark.
PACS contains four domains: \textbf{P}hoto, \textbf{A}rt painting, \textbf{C}artoon, and \textbf{S}ketch. 
It has a total of 7 categories and 9991 images. 
The size of each image is 227 $\times$ 227. 
We followed the official divide of training, validation, and testing set for a fair comparison.

\textbf{Implementation details.} 
In comparison with other SDG methods, we follow their settings.
We use the images in the Photo domain as the training and validation set, and the model is tested on all the samples of each other domains: art painting (A), cartoon (C), and sketch (S). 
The average accuracy is also calculated for comparison. 
Fast Fourier transform is utilized to transform the image into the frequency domain.
Six Gaussian bandpass filters are used to decompose the image into six frequency slices. Their center frequencies and bandwidths are (0,6), (7,8), (20,20), (40,20), (60,20), and (92,44), respectively. 
This division is an empirical value to ensure that each frequency component contains certain effective information.
At the same time, six Gaussian bandstop filters are used in the stop branch, and their center frequency and bandwidth are consistent with those of the bandpass filters. During training, two corresponding frequency slices will be input to different branches of the framework for training at the same time. 
In addition to the frequency slices, the original image is input to both branches for learning the context information between frequency bands.
We use ResNet18 \cite{he2016deep} as the backbone, and the pre-trained parameters on ImageNet1K \cite{deng2009imagenet} provided by PyTorch \cite{paszke2019pytorch} are used. 
The shape of each branch output feature is 512. 
And the final used linear layer size is input 1024 and output 7. 
We use Adam \cite{kingma2014adam} optimizer. 
The initial learning rate is 0.0001 and adjusted by the cosine annealing algorithm. 
The weight decay is 0 by default. 
Unless otherwise stated, the epoch is set to 100, and the batch size is set to 32. 
We tested the influence of different values of the hyper-parameter $\alpha$ on the generalization performance and took the $\alpha = 5$ corresponding to the highest precision as the final hyper-parameter. 
All experiments are implemented with PyTorch and run on an NVIDIA Tesla V100 GPU.

\subsection{Effective Frequency Slices Are Not Learned Adequately.}
\label{exp:freq}

\textbf{Frequency slices in Photo domain.}
We assume that for a deep learning model trained under an experienced risk minimization (ERM) strategy, some frequency components of images that may have a positive effect on image classification may not be learned adequately. 

\begin{figure}[b]
	\centering          
	\includegraphics[scale=1.]{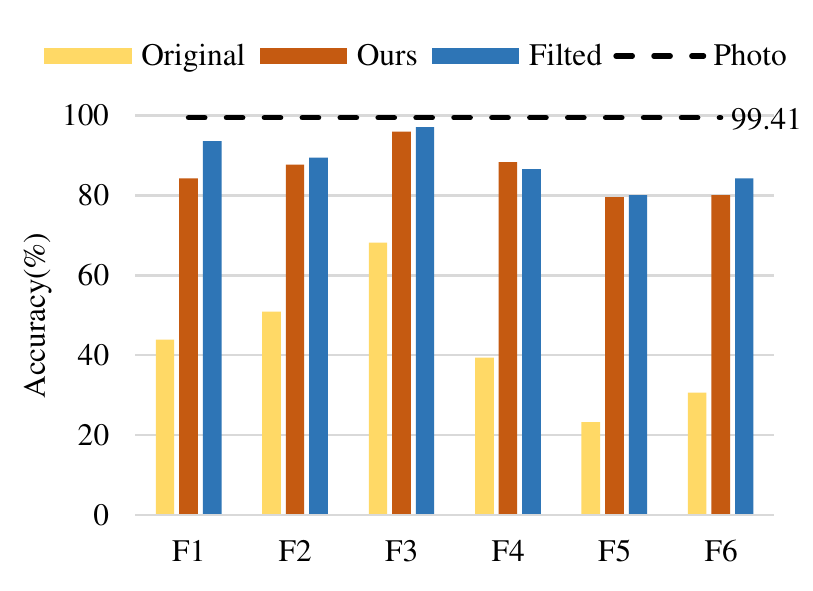}  
\caption{Accuracy(\%) histogram of 3 different training settings. The testing is conducted on the frequency slices. The yellow bars represent training on the original photo samples, the red bars represent our method, and the blue bars represent trained separately on each frequency component. F1 $\sim$ F6 means testing on each frequency component with F1 lowest and F6 highest. The dotted line is trained and tested on original photo images.}
\label{fig:useful}
\end{figure}

To verify this idea, we first designed an experiment to observe whether each frequency component of the image is effective and whether it has been learned. 
And the results can be found in \cref{fig:useful}. 
Specifically, we split the photo images into six frequency slices (F1$ \sim $F6) according to \cref{eq:FTpass}, where F1 represents the lowest frequency, and F6 is the highest frequency band. 
We have designed three training settings. 
Firstly, we trained a DNN on the original photo images and then tested it on the frequency slices to explore whether the model could learn each frequency component. 
Higher accuracy on each frequency slice indicates more adequately the frequency component is learned. 
We also tested the model with original photo images as a contrast. 
Then we trained and tested the model on each separated frequency slice to explore whether each frequency component was effective for the classification task. 
Higher accuracy on each frequency slice indicates the more effective the frequency component is. 
Finally, the proposed method was used to verify whether it is helpful for learning various frequency components.

Based on the results shown in \cref{fig:useful} there are some observations.

\begin{itemize}
    \item Compared with the testing accuracy on the original images (dotted line), models trained on original images can hardly achieve comparable accuracy on separated frequency slices (yellow bars). It reveals that even if the model has achieved high classification accuracy, it did not learn all frequency components adequately. Besides, the phenomenon is more obvious for the high-frequency component. We think the reason is related to CNN's preference to learn low-frequency components first \cite{xu2019frequency}, and also because most of the energy in the image is concentrated in low-frequency \cite{yang2020fda}. So the high-frequency component is more difficult to be concerned about. 
    \item The accuracy of training with each separated frequency slice (blue bar) is much higher and very close to the results on the original image (dotted line). It is strong evidence that each frequency slice contains a considerable number of effective information for the classification task. 
    \item Model trained with our approach (red bar) achieves a performance close to training with separated frequency slices (blue bar) for each frequency band. By comparing the three groups of bars, it can be concluded that our method significantly improves the model's learning of effective information in all the frequency bands.

\end{itemize}

\textbf{Different domains do not share similar main frequency bands.}
According to the last experiment, we found that for a single domain, all the frequency slices contain effective information for the classification task, but they are not fully learned.
Furthermore, we propose two extension problems, which will affect the design idea of the method.
(1) Is this a special case of some domains?
(2) For different domains, whether the distributions of effective information in the frequency bands are similar. 

We extend the experiment to all four domains of the PACS dataset, and the experimental results are shown in \cref{tab:domainfreq}.
It is trained and tested with each separated frequency band. 
Higher accuracy indicates more effective information for classification is contained in the frequency band.

\begin{table}[t]
  \begin{center}
  \caption{Accuracy(\%) of different domain frequency slices. They are all trained with a specific frequency band of a domain and tested with the samples in the same distribution of themselves. The best result of each domain is in \textbf{bold faces}. In comparison, the accuracy of random guess is 14.29\%.}
  \label{tab:domainfreq}
  \resizebox{0.47\textwidth}{!}{
    \begin{tabular}{l|c c c c c c} 
      \hline
      \textbf{Domain} & \textbf{F1} & \textbf{F2} &\textbf{F3} &\textbf{F4} &\textbf{F5} &\textbf{F6}\\
      \hline
      Photo   &93.57	&89.47	&\textbf{97.08}	&86.55	&80.12	&84.21 \\ 
      Art painting   &77.2 & 76.17 & \textbf{83.42} & 74.61 & 63.73 & 63.73 \\ 
      Cartoon   &\textbf{96.61} & 90.68 & 93.64 & 91.53 & 82.63 & 84.75  \\ 
      Sketch      &90.7 & 92.96 & \textbf{95.98} & 94.72 & 93.22 & 94.97 \\ 
      \hline
    \end{tabular}}
  \end{center}
\end{table}

According to the experimental results, the answers to the previous two questions can be given.
(1) It is not a special case that the effective information of each frequency band has not been fully learned. It exists widely in all the tested domains.
(2) Different domains do not share similar main frequency bands. For example, Cartoon has the most effective information in the lowest frequency band because the image in this domain is represented by a lot of colors which are related to low-frequency. Meanwhile, the main frequency band of Sketch has a higher frequency because lines are more high-frequency than gradient colors. 
Therefore, it is important to fully learn the effective information of each frequency band for domain generalization tasks.
Since the effective information distribution of each frequency band is different for domains, it is more reasonable to learn on different frequency slices than simply assign weights to frequency bands.

\subsection{Evaluation of Single Domain Generalization}

\begin{table}[b]
  \begin{center}
    \caption{SDG accuracy(\%) on PACS. One domain is used to train the model, and other domains are used for testing. And the accuracy is the average of the accuracy of the three domains.
    The best results are in \textbf{bold faces}, and the second best results are \underline{underlined}. }
    \label{tab:maincompareave}
    \begin{tabular}{l|c c c c|c} 
      \hline
      \textbf{Method} & \textbf{P} &\textbf{A} & \textbf{C} &\textbf{S} &\textbf{Avg.} \\
      \hline
      ERM \cite{volpi2018generalizing}       & 42.2	&70.9	&76.5	&53.10 & 60.7\\	
      MixStyle \cite{zhou2021mixstyle}     & 41.2 & 61.9 & 71.5  & 32.2 & 51.7 \\
      EFDMix  \cite{zhang2022exact}      &42.5 & 63.2  & 73.9  & 38.1 & 54.4 \\
      RSC \cite{huang2020self}       & 41.63 &70.67  & 75.08 & 47.25 & 58.66\\
      SelfReg \cite{kim2021selfreg}          & 43.46 & 72.59 & 76.56 & 45.76 & 59.59 \\
      L2D \cite{wang2021learning}       & 52.29 & \underline{76.91} & \underline{77.88} & 53.66 & 65.18 \\
      ASR \cite{fan2021adversarially}       & \underline{54.6} & 76.7 & \textbf{79.3} & \textbf{61.6} & \underline{68.1} \\
      \hline
      ours   & \textbf{64.52} & \textbf{79.91} &  77.63 & \underline{57.68} & \textbf{68.94} \\
      \hline
    \end{tabular}
  \end{center}
\end{table}

PACS is a widely used dataset for SDG tasks. 
We follow the most common strategy, leave-one-out, to test the SDG performance of the proposed approach.
Specifically, each domain is used as the source domain, in turn. Meanwhile, other domains are used as the target domain. Then the average result of a group of experiments is calculated to measure the SDG capability.
The results are shown in \cref{tab:maincompareave}.

Furthermore, we show the accuracy tested on other domains when \textbf{P}hoto domain, as the most common domain, is used as the source domain.
And the results are shown in \cref{tab:maincompare}.

\textbf{Single domain generalization on PACS.} 
We trained our method with each domain in the PACS dataset. 
The domain name in \cref{tab:maincompareave} represents the source domain, and the result is an average of the accuracy on the corresponding three target domains.

It can be seen that our approaches generally outperform other methods. 
An interesting observation is that compared with the previous method, our method performs particularly well on P, while the improvement on S is relatively small.
A possible reason is that there may be some similarity between the effective information of frequency slices. 
For the Photo domain, the difference between frequency slices is larger. Thus, our approaches can learn more effective information. In contrast, the difference between frequency slices is smaller for the Sketch domain.  

\textbf{Performance on different target domains.}
\Cref{tab:maincompare} shows the evaluation of P$\rightarrow$ACS. Our method achieved state-of-the-art results on this challenging benchmark. The result shows that our method can greatly enhance the generalization capability of the model in testing, which is due to better learning of the effective features of each frequency.

\begin{table}[t]
  \begin{center}
    \caption{SDG accuracy(\%) on PACS. Models are trained on photo and test on other domains (\ie art painting, cartoon, sketch). The best results are in \textbf{bold faces} and the second best results are \underline{underlined}. }
    \label{tab:maincompare}
    \begin{tabular}{l|c c c|c} 
      \hline
      \textbf{Method} & \textbf{A} & \textbf{C} &\textbf{S} &\textbf{Avg.}\\
      \hline
      ERM \cite{volpi2018generalizing}       & 54.43	&42.74	&42.02	&46.39\\	
      JiGen \cite{carlucci2019domain}     & 54.98 & 42.62 & 40.62 & 46.07\\
      RSC \cite{huang2020self}       & 56.26 &39.59  & 47.13 & 47.66 \\
      ADA \cite{volpi2018generalizing}       & 58.72 & 45.58 & 48.26 & 50.85 \\
      M-ADA \cite{qiao2020learning}     &\underline{58.96}	&44.09	& 49.96 &51.00 \\
      L2D \cite{wang2021learning}       & 56.26 & 51.04 & 58.42 & 55.24 \\
      MetaCNN \cite{wan2022meta} 	&54.05	&\textbf{53.58}  & \underline{63.88}	&\underline{57.17} \\
      \hline
      ours      & \textbf{66.41} & \underline{53.07} &  \textbf{74.10}& \textbf{64.52} \\   \hline
    \end{tabular}
  \end{center}
\end{table}
The improvement in Sketch is particularly obvious, which is over 10\% higher than MetaCNN, the last state-of-the-art method. The reason might be that sketches are made of lines without color, so most of the effective information is contained in the high-frequency component, which is easily ignored in the training process of general models. Our method has solved this problem well and achieved good results.

\subsection{Ablations}
The proposed framework contains two branches, and the similarity between their outputs is calculated as a loss to assist in learning the effective information in the two frequency domains.
To verify the effectiveness of each part of the framework, we conducted ablation experiments, and the results can be found in \cref{tab:branch}.
\begin{table}[t]
  \begin{center}
  \caption{Accuracy(\%) of models trained on single branch or without $L_{cons}$. Photo is the source domain, and A, C, and S are the target domains.}
    \label{tab:branch}
    \begin{tabular}{l|c c c |c} 
      \hline
       Method  & \textbf{A} & \textbf{C} &\textbf{S} &\textbf{Avg.}\\
      \hline
      pass branch   &56.87   &51.11    &68.35   &58.42\\
      stop branch &58.72   &41.60    &55.37   &51.89\\
      two same branches & 66.36 &29.39 &33.22   & 42.99\\
      ours w/o $L_{cons}$  &64.16   &41.64    &58.34   &54.71\\
      ours &  66.41 & 53.07 &  74.10 & 64.52 \\
      \hline
    \end{tabular}
  \end{center}
\end{table}


\textbf{About two branches.}
There is no structural difference between the two branches. 
If the branch is taken out separately, the difference between the two is that they receive complementary frequency slices as input.
Both of the single-branch experiment outperforms ERM.
We also tested two-branch ResNet18 as a control experiment. 
And the results are also better than ERM but not comparable to ours.

An interesting observation is that the accuracy of the pass branch is higher than the stop branch. 
The only difference between the two single branches is that the samples in the pass branch have narrower frequency bands. 
So a reasonable explanation is that the narrower frequency bands help the model concentrates on the specific frequency and not be disturbed by other information. So the model can learn the features of each frequency component and extract domain-invariant features.

\textbf{About the consistency loss.}
Then the consistency loss is removed, and significant performance degradation can be observed.
The two features for calculating consistency loss represent two complementary frequency components of an image. 
Since they are supposed to express information about the same object, they should be similar. 

The existence of consistency loss helps the framework learn features related to classification tasks, rather than the confusion by interference information related to the frequency domain or source domain. 
Thus, the classification accuracy is improved in the domain generalization task with the help of consistency loss between the feature extracted from the two branches.

\subsection{Sensitivity of Hyper-parameters}
The proposed method also introduces some hyper-parameters, and the adjustment of these hyper-parameters is not complicated. 
We conduct further analysis through the following experiments.

\begin{figure}
	\centering          
	\includegraphics[scale=1.0]{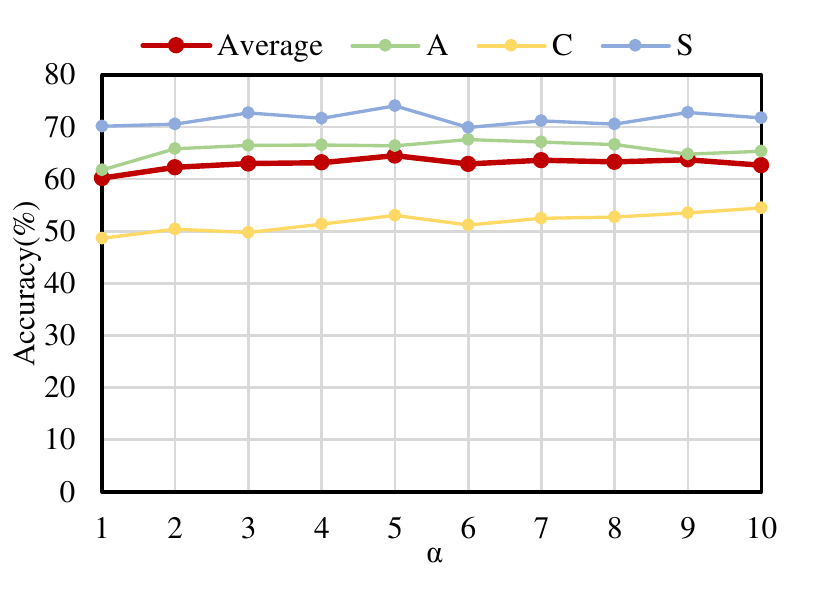}  
\caption{Accuracy of the model on the test domains with the variation of $\alpha$ from 1 to 10. The red line is the average of the other three lines.}
\label{fig:alpha}
\end{figure}

\begin{table}[b] 
  \begin{center}
  \caption{Accuracy(\%) of models trained on original samples and tested on every frequency slide. Low accuracy in each clone proves that many frequencies are not learned well.}
  \label{tab:n_slices}
    \begin{tabular}{l|c c c |c} 
      \hline
      K & \textbf{A} & \textbf{C} &\textbf{S} &\textbf{Avg.}\\
      \hline
      2   & 63.38	&44.71	&66.00     &58.03 \\
      4   & 63.92	&54.05	&68.77  & 62.25\\
      6   & 66.41	&53.07	&74.10 &64.52 \\	
      8   & 66.65	&48.36	&68.94 &61.32 \\
      \hline
    \end{tabular}
  \end{center}
\end{table}

\textbf{About $\alpha$.} 
We tried different $\alpha$ values, and the results are shown in \cref{fig:alpha}. 
In general, the performance of the model is not sensitive to the weight of $\alpha$.
For all tested $\alpha$, the average performance of the model is always above 60\%, which is still significantly higher than the previous methods.
A closer observation of the experimental results shows that with the increase of $\alpha$, the accuracy increases at first. 
It indicates that urging the network to extract consistent features of different frequencies can improve the effectiveness of features. 
Meanwhile, with the further increase of $\alpha$, the accuracy decreases gradually because the optimization of classification loss is affected.
Finally, we selected the hyper-parameter value with the highest accuracy, namely $\alpha$ = 5.

\textbf{About the number of frequency slices K.} 
Another hyper-parameter is related to the division of frequency bands. 
We tried to decompose the image into a different number of slices and carried out experiments. The results are shown in \cref{tab:n_slices}.
All the tested decomposition methods can achieve better performance than previous methods. 
Among them, decomposing the image into 6 slices is the best.
The too rough or fine division will have a certain impact on the performance.
The too rough division will make the model unable to fully learn the effective information in each slice, while too fine division will cause too little effective information in each slice, thus increasing the difficulty of learning.
For most image classification tasks, we think 6$\sim$8 are more appropriate. Furthermore, some automatic partitioning methods could be an improvement direction.

\subsection{Visualization of Features}
\label{subsec:tsne}
To further demonstrate the effectiveness of our approach, we use t-SNE \cite{van2008visualizing} to visualize the distribution of the unseen target features in the sketch and the art painting domain. We train two models with the ERM method and our approach on the Photo domain and test them on the Sketch and art painting samples separately, in which our approach has the largest and smallest improvement. We use the first 150 samples of each category to the plot.

\begin{figure}[htbp]
\centering

  \begin{subfigure}{0.45\linewidth}
  \centering
    \includegraphics[width=\linewidth]{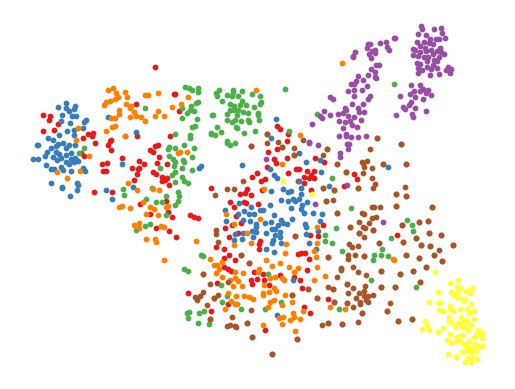}
    \caption{ERM, P$\rightarrow$S }
    \label{fig:short-a}
  \end{subfigure}
  \begin{subfigure}{0.45\linewidth}
  \centering
    \includegraphics[width=\linewidth]{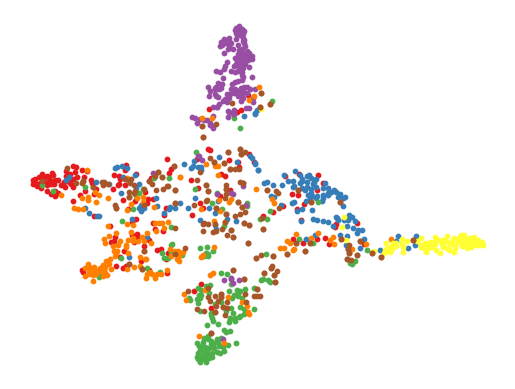}
    \caption{Ours, P$\rightarrow$S}
    \label{fig:short-b}
  \end{subfigure}
  
  \begin{subfigure}{0.45\linewidth}
  \centering
    \includegraphics[width=\linewidth]{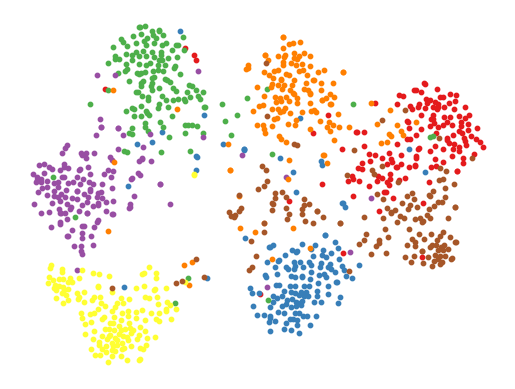}
    \caption{ERM, P$\rightarrow$A}
    \label{fig:short-c}
  \end{subfigure}
  \begin{subfigure}{0.45\linewidth}
  \centering
    \includegraphics[width=\linewidth]{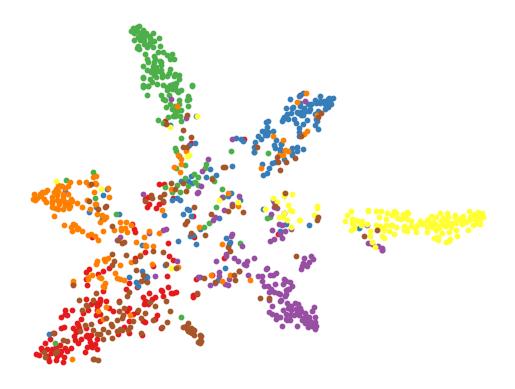}
    \caption{Ours, P$\rightarrow$A}
    \label{fig:short-d}
  \end{subfigure}

\centering
\caption{The t-SNE visualizations of target feature distribution for ERM and our approach. The models are trained on Photo and tested on Sketch and Art painting. Features with the same semantic label are drawn in the same color.}
\label{fig:tsne}
\end{figure}

It can be seen in \cref{fig:tsne} that the same category features extracted by our approach are gathered more tightly than by the ERM method. 
And features of different categories are more distant by our method.
Our approach obviously has better class separation than the baseline model, which indicates that our method extracts efficient classification features in different unseen domains.

\section{Conclusion}
In this paper, we first share an observation that each frequency component contains effective information that cannot be ignored, but not all of them are learned by deep learning models. 
Based on the observation, we proposed a new domain generalization method, which can better learn the effective features contained in each frequency band.
Through frequency decomposing, our approach can better learn efficient features from all the frequency bands.
Sufficient experiments from multiple angles indicate that our approach outperforms the state-of-the-art single domain generalization methods.



{\small
\bibliographystyle{ieee_fullname}

}

\end{document}


\title{Supplementary Materials for Frequency Decomposition to Tap the Potential of Single Domain for Generalization}
\maketitle

\thispagestyle{empty}

\appendix
\section{Example of Frequency Slices}

\cref{fig:apd_1} shows the original images and the frequency slices of PACS. It can be seen in the first column that the style of different domains is very different. And different column shows different frequency bands. The low-frequency band has more colors and energy and the high-frequency band has lines and less energy except for Sketch. 

\begin{figure}[htbp]
\centering
\includegraphics[width=1\linewidth]{cvpr2023-author_kit-v1_1-1/latex/appendix_fig/freqexample.pdf}
\caption{PACS image frequency slices of six frequency bands. The four rows are photo, art painting, cartoon, and sketch. The seven columns are original images and F1$\sim$F6 frequency slices.}
\label{fig:apd_1}
\end{figure}

\section{Accuracy of Different Frequency Slices} 

\Cref{tab:frequency_acc} is the accuracy value of three training settings consistent with Fig. 3 in the main paper. From the value, we can see the performance gap of the three methods in each frequency band. The accuracy of the model trained on the original image is 28.89\% $\sim$ 56.84\% lower than that trained on each frequency slice, which is a very big gap. It indicates that training on the original image can not learn the information of each frequency component well, while there is effective information in every frequency band indeed. The accuracy of our method is very close to that of training on each frequency band, which shows that our method has learned effective information in each frequency band well.

\begin{table}[htbp]
  \begin{center}
  \caption{Accuracy(\%) of different training methods with testing on the frequency slices. The first two lines are trained on the original photo samples and on each frequency component of photo images. The third line is the method proposed by us.  Each column means a frequency component for test with F1 lowest and F6 highest.}
  \label{tab:frequency_acc}
  \resizebox{0.47\textwidth}{!}{
    \begin{tabular}{l|c c c c c c} 
      \hline
      & \textbf{F1} & \textbf{F2} &\textbf{F3} &\textbf{F4} &\textbf{F5} &\textbf{F6}\\
      \hline
      Original   &43.86	&50.88	&68.19	&39.41	&23.28	&30.64 \\ 
      Filtered   &93.57	&89.47	&97.08	&86.55	&80.12	&84.21\\ 
      Ours      &84.21	&87.72	&95.91	&88.3	&79.53	&80.12 \\
      \hline
    \end{tabular}}
  \end{center}
\end{table}

\section{Similar Information Between Frequency Slices}

\begin{table}[htbp]
  \begin{center}
  \caption{Accuracy(\%) of photo domain slices. They are trained with the left column and tested with the top row. The best accuracy of each row is in \textbf{bold faces}.}
  \label{tab:freqcrossp}
  \resizebox{0.48\textwidth}{!}{
    \begin{tabular}{c|c c c c c c} 
      \hline
      \textbf{Photo} & \textbf{F1} & \textbf{F2} &\textbf{F3} &\textbf{F4} &\textbf{F5} &\textbf{F6}\\
      \hline
        \textbf{F1} & \textbf{91.22} & 29.82 & 30.99 & 23.39 & 19.3 & 18.13 \\
        \textbf{F2} & 30.41 & \textbf{88.89} & 78.36 & 35.67 & 19.88 & 26.32 \\
        \textbf{F3} & 14.04 & 54.39 & \textbf{95.91} & 72.51 & 26.32 & 40.94 \\ 
        \textbf{F4} & 23.98 & 35.67 & 87.13 & \textbf{88.89} & 75.44 & 79.53 \\ 
        \textbf{F5} & 21.05 & 21.05 & 45.03 & 77.78 & \textbf{83.63} & 78.36 \\ 
        \textbf{F6} & 19.88 & 16.37 & 32.75 & 73.1 & 79.53 & \textbf{83.63} \\ 
      \hline
    \end{tabular}}
  \end{center}
\end{table}
\begin{table}[ht]
    \centering
    \caption{Accuracy(\%) of sketch domain slices. They are trained with the left column and tested with the top row. The best accuracy of each row is in \textbf{bold faces}.}
    \label{tab:freqcrosss}
    \resizebox{0.48\textwidth}{!}{
    \begin{tabular}{c|c c c c c c}
    \hline
        \textbf{Sketch} & \textbf{F1} & \textbf{F2} &\textbf{F3} &\textbf{F4} &\textbf{F5} &\textbf{F6} \\ \hline
        \textbf{F1} & \textbf{90.7} & 50.75 & 56.28 & 57.29 & 27.64 & 31.66 \\ 
        \textbf{F2} & 26.38 & \textbf{92.71} & 85.43 & 85.18 & 52.51 & 54.78 \\ 
        \textbf{F3} & 24.37 & 83.17 & \textbf{96.23} & 89.95 & 63.32 & 74.37 \\ 
        \textbf{F4} & 34.17 & 82.16 & 85.23 & \textbf{95.23} & 76.88 & 77.64 \\ 
        \textbf{F5} & 19.10 & 57.54 & 85.68 & \textbf{94.72} & \textbf{94.72} & 81.91 \\ 
        \textbf{F6} & 19.10 & 71.11 & 82.16 & 87.19 & 86.43 & \textbf{93.22} \\ 
        \hline
    \end{tabular}}
\end{table}

Different frequency bands may contain similar information, which we called cross-information.
The proportion of cross information between different frequency bands of different domains may be different.
For example, Photo has bright colors and borders, while Sketch only has object outlines without colors. We tested in these two extreme cases. 
Specifically, we train the model in each frequency band and test it with all frequency bands. The higher accuracy means the information contained in the two frequency bands is more similar, and more cross-information.

Based on the results shown in \cref{tab:freqcrossp} and \cref{tab:freqcrosss}, there is more similar information in closer frequency bands. The accuracy of photo between frequency bands is almost below 80\%, while many of them in the sketch domain is more than 80\%. There is less cross-information between various frequency bands of Photo than Sketch. This is maybe a reason why the performance of training on Photo is better than on Sketch. With less cross-information, the model can learn more information with less disturbance in each frequency band and achieve better performance.

\section{Visualization of Frequency Band 3 Features}

We use t-SNE to visualize the distribution of the F3 slices features of the photo domain, associate with column F3 in \cref{tab:frequency_acc}. The models of (a) and (b) are trained on the original photo samples and on frequency slices of F3. The model of (c) is the method proposed by us. 
\Cref{fig:tsne_F3} shows the result. It can be seen that the ERM method cannot distinguish various categories well, while training on the F3 and our method can extract features with good classification. This is consistent with the accuracy in \cref{tab:frequency_acc}.

\begin{figure}[htbp]
\centering
  \begin{subfigure}{0.48\linewidth}
  \centering
    \includegraphics[width=\linewidth]{cvpr2023-author_kit-v1_1-1/latex/appendix_fig/ermP_F3.png}
    \caption{Trained by ERM}
    \label{fig:F3erm}
  \end{subfigure}
  \begin{subfigure}{0.48\linewidth}
  \centering
    \includegraphics[width=\linewidth]{cvpr2023-author_kit-v1_1-1/latex/appendix_fig/F3P_F3.png}
    \caption{Trained with F3}
    \label{fig:F3}
  \end{subfigure}
  \begin{subfigure}{0.48\linewidth}
  \centering
    \includegraphics[width=\linewidth]{cvpr2023-author_kit-v1_1-1/latex/appendix_fig/oursP_F3.png}
    \caption{Trained by ours}
    \label{fig:F3ours}
  \end{subfigure}
\centering
\caption{The t-SNE visualizations of Photo F3 frequency slices feature distribution, associated with column F3 in \cref{tab:frequency_acc}. (a) The model is trained on the original photo images. (b) The model is trained on F3 frequency slices. (c) The model is trained on Photo with our approach. Features with the same semantic label are drawn in the same color.}
\label{fig:tsne_F3}
\end{figure}

\section{Broader Impact}
In this paper, we provide a solution for single domain generalization, which enables the model to achieve a better generalization effect with a single domain training set. It improves the adaptability of the model and reduces the cost of collecting multi-source data. Meanwhile, compared with the method of generating new domain images, our approach of mining the information of the data itself eliminates the possibility of introducing unrealistic information or offensive content. According to our knowledge, our work may not adversely affect the moral aspects and future social consequences.